# Time Complexity Analysis of Binary Space Partitioning Scheme for Image Compression

Rehna. V. J, M. K. Jeyakumar

*Abstract*— Segmentation-based image coding methods provide high compression ratios when compared with traditional image coding approaches like the transform and sub band coding for low bit-rate compression applications. In this paper, a segmentation-based image coding method, namely the Binary Space Partition scheme, that divides the desired image using a recursive procedure for coding is presented. The BSP approach partitions the desired image recursively by using bisecting lines, selected from a collection of discrete optional lines, in a hierarchical manner. This partitioning procedure generates a binary tree, which is referred to as the BSP-tree representation of the desired image. The algorithm is extremely complex in computation and has high execution time. The time complexity of the BSP scheme is explored in this work.

## I. INTRODUCTION

The field of image compression is rich in diverse source coding schemes ranging from classical lossless techniques and popular transform approaches to the more recent segmentation-based (or second generation) coding methods [1]. For low bit-rate compression applications, segmentation-based coding methods provide, in general, high compression ratios when compared with traditional transform, vector quantization (VQ), and sub band (SB) coding approaches. In the past decades, the discrete cosine transform (DCT) has been the most popular for compression because it provides optimal performance and can be implemented at a reasonable cost. Several compression algorithms, such as the JPEG standard for still images and the MPEG standard for video images are based on DCT [12]. However, the EZW, the SPIHT, the SPECK, the EBCOT algorithms and the current JPEG 2000 standard are based on the discrete wavelet transform (DWT). DWT has the ability to solve the blocking effect introduced by DCT, it also reduces the correlation between the neighboring pixels and gives multi scale sparse representation of the image. In spite of providing excellent results in terms of rate-distortion compression, the transform-based coding methods do not take an advantage of the underlying geometry of the edge singularities in an image. The second generation image coding techniques [1] exploit the geometry of the edge singularities of an image. The first segmentation-based coding methods appeared in the early 1980s. These algorithms partition the image into geometric regions over which it is approximated using low-order polynomials. Many variations have since been introduced and among them, the binary space partition (BSP) scheme [2] is a simple and effective one. BSP-tree-based image coding system that is capable of achieving high compression ratios when applied on still images. The most challenging aspect of a segmentation based coding approach is to balance between a small number of geometrically simple regions and the smoothness of the image signal within these regions. BSP scheme achieves the above balance by using a simple, yet flexible description of the images [3]. It has wide applications in the field of image processing and computer graphics. But the main drawback of this technique is its high execution time that makes it less practically applicable. The present study is envisaged to analyze the time complexity of the BSP algorithm. This method is applied to 8 bits grayscale images but it could be extended to color images in the same way that JPG2000 has been applied to different type of images (i.e. 8 bits/pixel, 24 bits/pixel). The rest of the paper is organized as follows: In section II, preliminaries on image compression are discussed. Section III and its subsections gives an overview of the algorithm for binary tree generations of images, the partitioning method (based on this work and other related work), quantization, tree encoding and decoding are provided. The implementation of the algorithm is dealt in section IV followed by results and discussion in section V and the paper is concluded in section VI.

## II. IMAGE COMPRESSION PRELIMINARIES

Image Compression aims to reduce the number of bits required to represent an image by removing the redundancies which makes it one of the most useful and commercially successful technologies in the field of Digital Image Processing. Three principle types of data redundancies that can be identified are:

*Coding redundancy:* Coding redundancy consists of variable length code words selected as to match the statistics of the original source. In the case of Digital Image Processing, it is the image itself or the processed version of its pixel values. Examples of image coding schemes that explore coding redundancy are the Huffman codes and the Arithmetic coding technique.

*Spatial redundancy:* Spatial redundancy is sometimes called interframe redundancy, geometric redundancy or interpixel redundancy. Here, because the pixels of most 2-D intensity arrays are correlated spatially that is each pixel is similar to or independent on neighbouring pixels, information is unnecessarily replicated in the representation of the correlated pixels. Examples of this type of redundancy include Constant area coding and many Predictive coding algorithms.

*Irrelevant information:* Most 2-D intensity arrays contain information that is ignored by the human visual system. Image and video compression techniques aim at eliminating or reducing any amount of data that is psycho visually redundant. Most of the image coding algorithms in use today exploit this type of redundancy, such as the discrete cosine transform based algorithm at the heart of the JPEG encoding standard. The usual steps involved in compressing an image are, specifying the rate and distortion parameters for the target image, dividing this image data into various classes, based on their importance, then dividing the available bit budget among these classes, such that the distortion is minimum. Next step involves quantizing each class separately using the bit allocation information followed by encoding of each class using an entropy coder and write to the file.

### III. ALGORITHM DESCRIPTION

#### A. BSP Tree Generation

The basic concepts of the BSP technique are described here. Given an image f in $\Omega \subset R_2$ and $\Omega = [0; 1]^2$, the algorithm subdivides $\Omega$ into two subsets $\Omega_0$ and $\Omega_1$ using a bisecting line and minimizing a given functional [4]. The algorithm continues partitioning each region recursively until it reaches a given measure or there are no enough pixels to subdivide. The algorithm constructs a binary tree with the partitioning information. To approximate the image f at any region $\Omega_i$, they use two bivariate linear polynomials defined by:

$$Q_{\Omega_i} = A_i\,x + B_i\,y + C_i. \quad (1)$$

The functional used to find the best subdivision for a given region is the following:

$$F(\Omega_0, \Omega_1) = \arg\min_{\Omega_0,\Omega_1} \|f - Q_{\Omega_0}\|^2_{\Omega_0} + \|f - Q_{\Omega_1}\|^2_{\Omega_1} \quad (2)$$

Where $\Omega_0$ and $\Omega_1$ represent the subsets resulting from the subdivision of $\Omega$, with the constraints $\Omega = \Omega_0 \cup \Omega_1$ and $\Omega_0 \cap \Omega_1 = \emptyset$; $\Omega_0$ and $\Omega_1$ are consider as children for a given region $\Omega$ which is called the father. Figure.1 shows the steps of the BSP algorithm, first $\Omega$ is subdivided by L into two subdomains $\Omega_0$ and $\Omega_1$, then $\Omega_0$ is subdivided by $L_0$ into $\Omega_{00}$ and $\Omega_{01}$. It continues to subdivide until either the area of the domain is too small, ie, it contains only a few pixels or the approximation error is sufficiently small. It is possible to represent this as a tree structure as is shown in Figure 2.

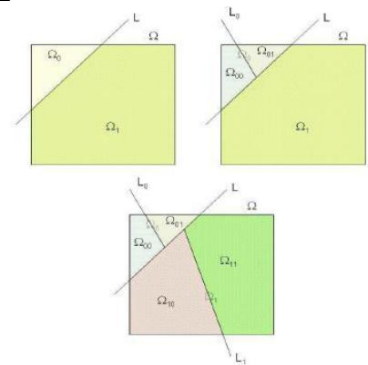

Fig. 1 Two Partition Levels Using Bisecting Lines

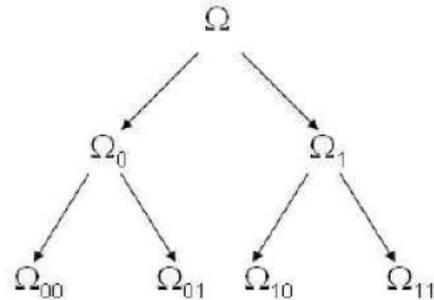

Fig. 2 BSP Tree Representation

The most critical aspect of the BSP-tree method is the criterion used to select the partitioning lines of the BSP tree representation. The optimal bisection is computed from a collection of discrete optional lines. The selected line of partition is the line that minimizes this equation. The bisecting lines of the BSP scheme in, are uniformly quantized using the normal representation of the straight line,

$$\rho = x \cos\theta + y \sin\theta \quad (3)$$

Where $\rho$ is the normal distance between the line and an origin point close to the subdivided domain and $\theta$ is the angle between the line's normal and axis. The set of possible bisecting lines in [7], [8] is discritized by the parameters $\theta$ and $\rho$ of the normal representation of the line. Quantization step size of the parameter $\theta$ depends upon the size of the polygonal domain which is defined as the smallest power of two. Quantization step size of $\theta$ in [4] is $(3\pi/2.\#\theta)$ where $\#\theta$ is the number of line orientations and is given by:

$$\#\theta = \min_{2^j \geq \sqrt{M \cdot N}} 2^j \quad (4)$$

Where M and N are the dimensions of the subdivided domain area. The size of quantization step of parameter $\rho$ in [6] depends on the orientation $\theta$. Quantization step size of parameter $\rho$ corresponding to orientation $\theta_i$ is max($\cos\theta_i$, $\sin\theta_i$).

#### B. Quantization and Coding of Line Parameters

Once the bsptree is generated, the tree is pruned and appropriate encoding methods can be used to code the line data and the wavelet coefficients. Huffman coding [5] is used here for encoding. Quantization of the lines implies, in general, the quantization of the parameters used to represent these lines [6]. Here we focus on the normal representation (or the (θ, ρ) parameterization) since it simplifies the line detection process when compared with the slope intercept representation, y = mx+C. A point in the (θ, ρ) parameter space represents a straight line in the image space (2-D plane) [9]. From the dimensions of the image domain, one can exactly identify the set of all lines that can pass through (or partition) the domain. This set of partitioning lines corresponds to (or is represented by) a set of points in the parameter space. We refer to this set of points as the partitioning lines domain (PLD) [10]. The PLD domain is usually a connected region in the parameter space as shown in Fig. 3. The boundaries of the PLD domain of an image are defined by the parametric representation of the image-domain four corner points. For example, if the image domain has the corner points (z, y) = (1, 1), (-1, 1), (-1,-1), and (1,-1), then these points are mapped onto four sinusoidal functions in the parameter space using p = x cos θ + y sin θ. Now, for a given value of θ, the lower and upper limits of the PLD domain are determined by the minimum and maximum values of the four parametric (sinusoidal) curves generated by the four corner points of the image [11]. Similarly, knowing the geometric description of a given polygon, one can identify the parameter space domain of all possible lines that can pass through this polygon. In other words, each polygon in the image domain has a corresponding unique PLD domain in the parameter space.

### C. Decoding

The decoder will have a priori knowledge of the image dimensions (otherwise, the dimensions are sent first). Hence, the first partitioning line representation in the parameter space has to be within the PLD domain of the image. Due to the recursive nature of the BSP-tree representation, the BSP-tree decoder (at the receiver) has a complete geometric description of the polygon to be partitioned [8]. Consequently, the receiver has knowledge of the PLD domain of that polygon, and expects the parametric representation of the partitioning line to be within that PLD domain. As the recursive partitioning continues, the polygons get smaller and so do their corresponding PLD domains in the parameter space.

### IV. IMPLEMENTATION

The input image used for the implementation is the bitmap image of cameraman which is a 256x256 image. As mentioned the BSP tree generation procedure is computationally very intensive and takes a lot of time for execution. In order to reduce the computational complexity of the algorithm, the image is tiled and then the BSP algorithm is applied on each tile separately, thereby creating a BSP forest. Experiments show that the choice of tiles of size 128x128 significantly reduces the time complexity of the algorithm but does not reduce its coding efficiency. Tiling of cameraman image into 4 tiles with each tile having a size of 128x128 is shown in the Figure 3.

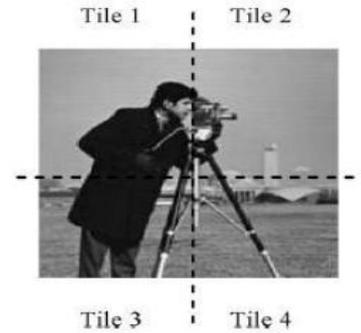

**Fig. 3 Image Tiling with Each Tile of Size 128 X 128**

Image tiling has one disadvantage that, at low bit-rates, there are blocking artifacts at the tiles' boundaries which is a phenomena similar to low bit-rate JPEG compression [12]. In addition, there is a possibility that we need to generate several BSP trees, one at each tile, whereas, with no tiling, only a single BSP is needed. In this implementation, the tile size is further reduced to 64x64 to reduce the time complexity which will divide the image into 16 parts as shown in Figure 4.

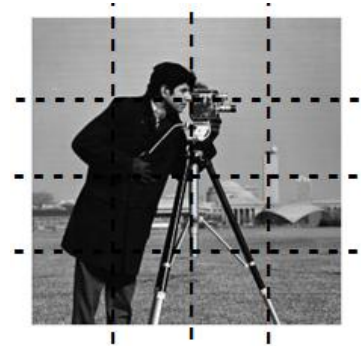

**Fig. 4 Image Tiling with Each Tile of Size 64 X 64**

Since the domain area is 64x64, the total number of pixels in this area would be 4096. Initially the partitioning limit considered is 2000 pixels. This means that, the image area would be partitioned recursively as long as there is 2000 or more number of pixels in the subarea. If the number of pixels in the end child polygon is less than 2000, partitioning procedure stops. Each recursive procedure of partitioning is an iteration. Now, there exists only 1 line, which is selected from a collection of discrete optimal lines according to the procedure mentioned earlier. This line divides the entire image area, making 2 partitions with the number of iterations equal to 3, one scanning the entire tile area, the other two for the 2 child polygonal areas. Further partitioning is not possible since each child polygonal area will have less than 2000 pixels. For a limit of 1000 pixels, the number of iterations required is 5, for a limit of 500 pixels; the number of iterations is 11 and so on. This is shown in Table 1. As the limit decreases, the number of partitions increases. This recursive procedure is terminated only when the area of the domain is too small, ie., it contains only a few pixels. In this work, the smallest domain area is considered to be consisting of 64 pixels. That is, the procedure of image partitioning stops when the number of pixels in the child polygon is less than 64.

## V. RESULTS AND DISCUSSION

The proposed algorithm is implemented using MATLAB and tested on the still image of Cameraman of sizes 256 x 256, respectively and bit depth 8. Each tile of the image is having a size of 64x64. Once the algorithm is implemented in this small area, it is extended to rest of the portions of the image. The image partitioning and BSP tree generation is the most intensive and time consuming part of the algorithm. Table I gives the time complexity of the BSP tree generation procedure. It shows the simulated results for the first tile of the input image using the computer specifications of Intel core i5 processor, 3GB DDR3 memory and a speed of 1.64 GHz. For a very small size of the image tile area, 64x64 and partitioning limit of 2000 pixels, with just 1 line drawn and number of iterations is 3, it takes almost 1½ hours for executing BSP tree generation algorithm. As the partitioning limit decreases, the number of iterations increases which in turn increases the number of lines drawn and consequently results in longer time for execution. We can see that for the partitioning limit of 64, the processing time was almost 25 hours. In the work done by Dekel et.al [3], the tile size was taken to be 128x128 which will take days or weeks for generation of BSP tree.

**Table I. Time Complexity Analysis of BSP Tree Generation For The First Tile Of The Input Image**

| Tile size | Partition Limit (no. of pixels) | No. of Iterations | Execution Time (seconds) | Execution Time (hours) |
|---|---|---|---|---|
| 64x64 | 2000 | 3 | 5844.95425 | 1.6 |
| 64x64 | 1000 | 5 | 14287.51548 | 3.9 |
| 64x64 | 500 | 11 | 22204.265208 | 6.16 |
| 64x64 | 250 | 23 | 38775.571362 | 10.7 |
| 64x64 | 128 | 51 | 66461.530830 | 18.4 |
| 64x64 | 64 | 107 | 93330.788558 | 25.3 |
| 128x128 | 8000 | 3 | 124739.257469 | 34.6 |
| 128x128 | 4000 | 5 | 203889.960401 | 56.6 |
| 128x128 | 2000 | 11 | 365760.200456 | 101.6 |

Once the bsptree is generated, the tree is pruned and encoding methods are used to code the line data and the wavelet coefficients. This process just takes a few seconds. The proposed method produces the PSNR values that are competitive with the state-of-art coders in literature like the SPIHT [13], EZW [14] etc. The comparative results with the SPIHT coder are shown in Table II.

**Table II. PSNR Values for Cameraman Test Image**

| Compression Ratio | SPIHT | BSP |
|---|---|---|
| 1:32 | 28 | 27.62 |
| 1:64 | 25 | 25.29 |
| 1:128 | 22.8 | 23.04 |

The results show that BSP based image compression provide high compression ratios with good PSNRs at low bit rates. Figure 5 shows reconstructed images of Cameraman using the proposed method at compression ratios of 64 and 128 respectively.

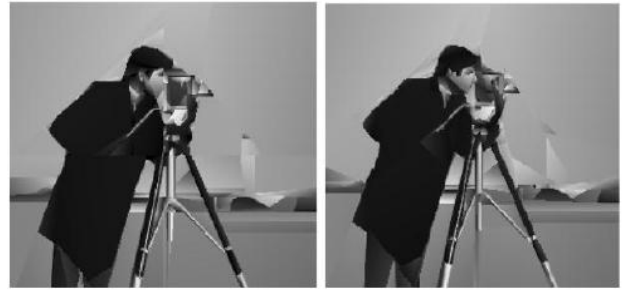

**Fig. 5 Reconstructed Cameraman Images at CR of 64 and 128 Using Proposed Method**

## VI. CONCLUSION AND FUTURE WORK

The paper presents the BSP based image coding algorithm. The algorithm is found to be extremely complex in computation and has high execution time. This makes the technique of image coding less practically applicable. The time complexity of the algorithm is analyzed here. In future, new methods to reduce the execution time may be explored. One of the prospective approaches is the introduction of soft computing techniques like neural networks.

## VII. ACKNOWLEDGMENT

Authors extend thanks to the Management and Principal, Noorul Islam Center for Higher Education, Kumara oil for their support and co-operation in carrying out the work successfully in the institute. The authors also thank the reviewers for their valuable comments and suggestions that helped us to make the paper in its present form.

**AUTHORS PROFILE**

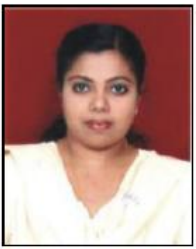

**Rehna. V. J** was born in Trivandrum, Kerala State, India in 1980. She studied Electronics & Communication Engineering at the PET Engineering college, Vallioor, Tirunelveli District, Tamilnadu State, India fom 1999 to 2003. She received Bachelor"s degree from Manonmanium Sundarnar University, Tirunelveli in 2003. She did post-graduation in Microwave and TV Engineering at the College of Engineering, Trivandrum and received the Master's degree from Kerala University, Kerala, India in 2005. Presently, she is a research scholar at the Department of Electronics and Communication Engineering, Noorul Islam Center for Higher Education, Noorul Islam University, Kumarakoil, Tamilnadu, India; working in the area of image processing under the supervision of Dr. M. K. Jeya Kumar. She is currently working as Assistant Professor at the Department of Electronics & Communication Engineering, HKBK College of Engineering, and Bangalore, India. She has served as faculty in various reputed Engineering colleges in South India over the past nine years. She has presented and published a number of papers in national/international journals/conferences. She is a member of the International Association of Computer Science & Information Technology (IACSIT) since 2009. Her research interests include numerical computation, soft computing, enhancement, coding and their applications in image processing.

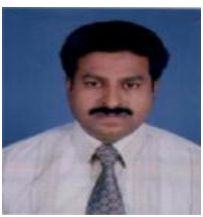

**Dr. M. K. Jeya Kumar** was born in Nagercoil, Tamilnadu, India on 18th September 1968. He received his Masters in Computer Applications degree from Bharathidasan University, Trichirappalli, and Tamilnadu, India in 1993. He fetched his M.Tech degree in Computer Science and Engineering from Manonmaniam Sundarnar University, Tirunelveli, and Tamilnadu, India in 2005. He completed his Ph.D degree in Computer Science and Engineering from Dr.M.G.R University, Chennai, and Tamilnadu, India in 2010.
He is working as a Professor in the Department of Computer Applications, Noorul Islam University, Kumaracoil, Tamilnadu, India since 1994. He has more than seventeen years of teaching experience in reputed Engineering colleges in India in the field of Computer Science and Applications. He has presented and published a number of papers in various national and international journals. His research interests include Mobile Ad Hoc Networks and network security, image processing and soft computing techniques.